\documentclass[runningheads]{llncs}
\usepackage{graphicx}
\usepackage{multirow}
\usepackage{todonotes}
\usepackage{comment}
\usepackage{adjustbox}
\usepackage{amsmath}
\usepackage[utf8]{inputenc}
\usepackage[english]{babel}
\usepackage{color}
\usepackage{amsfonts}
\usepackage{url}
\usepackage[mathscr]{euscript}
 \let\mathscr\relax%
\usepackage[scr]{rsfso}
\usepackage{caption}
\usepackage[labelfont=bf]{caption}
\usepackage{tabularx,booktabs}
\newcolumntype{C}{>{\centering\arraybackslash}X} 
\usepackage{multicol}
\usepackage{algorithm2e}
\usepackage{booktabs,ragged2e}
\usepackage[utf8]{inputenc}
\usepackage{todonotes}
\usepackage{lineno}
\usepackage{makecell}
\usepackage[flushleft]{threeparttable}
\usepackage{subcaption}
\usepackage{color}

\usepackage[nolist,nohyperlinks]{acronym}
\usepackage{footnote}
\usepackage[perpage]{footmisc}
\usepackage[autostyle]{csquotes}
\makesavenoteenv{tabular}
\acrodef{DL}{Deep learning}
\acrodef{CNN}{Convolutional Neural Network}
\acrodef{ML}{Machine Learning}
\acrodef{mIoU}{mean Intersection over Union}
\acrodef{GI}{gastrointestinal}
\acrodef{AI}{Artificial Intelligence} 
\acrodef{CADx}{computer aided diagnosis} 
\acrodef{CRC}{colorectal cancer}
\acrodef{DSC}{Dice Coefficient}
\acrodef{mDSC}{Dice Coefficient}
\acrodef{OOD}{Out-Of-Distribution}
\acrodef{SOTA}{State-of-the-art}
\acrodef{HD}{Hausdorff distance}

\begin{document}
\title{Prototype Learning for Out-of-Distribution Polyp Segmentation}
\titlerunning{Prototype Learning for Polyp Segmentation}
\author{ Nikhil Kumar Tomar, Debesh Jha,  Ulas Bagci}
\institute{Machine \& Hybrid Intelligence Lab, Department of Radiology, Northwestern University, Chicago, USA}

  \maketitle      
\begin{abstract}
Existing polyp segmentation models from colonoscopy images often fail to provide reliable segmentation results on datasets from different centers, limiting their applicability. Our objective in this study is to create a robust and well-generalized segmentation model named \textit{PrototypeLab} that can assist in polyp segmentation. To achieve this, we incorporate various violetlighting modes such as White light imaging (WLI), Blue light imaging (BLI), Linked color imaging (LCI), and Flexible spectral imaging color enhancement (FICE) into our new segmentation model, that learns to create prototypes for each class of object present in the images. These prototypes represent the characteristic features of the objects, such as their shape, texture, color. Our model is designed to perform effectively on out-of-distribution (OOD) datasets from multiple centers. We first generate a coarse mask that is used to learn prototypes for the main object class, which are then employed to generate the final segmentation mask. By using prototypes to represent the main class, our approach handles the variability present in the medical images and generalize well to new data since prototype capture the underlying distribution of the data. PrototypeLab offers a promising solution with a dice coefficient of $\geq$ 90\% and mIoU $\geq$ 85\% with a near real-time processing speed for polyp segmentation. It achieved superior performance on OOD datasets compared to 16 state-of-the-art image segmentation architectures, potentially improving clinical outcomes. Codes are available at https://github.com/xxxxx/PrototypeLab.


\end{abstract}
\section{Introduction}
The Cancer statistic 2023~\cite{siegel2023cancer} estimates that \acf{CRC} will be the third leading cause of cancer-related incidence and death in the United States. 
The United States Preventive Services Task Force (USPSTF) recommends \ac{CRC} screening at 45 years of age~\cite{ngcolorectal}. Thus, regular screening is essential as \ac{CRC} does not show symptoms (bleeding in the stool, constipation or diarrhoea) at an early stage. Studies have shown the lesion miss rate to be 26\%~\cite{corley2014adenoma}. This  emphasizes the need for an accurate and reliable screening \ac{CADx} method for reducing the polyp miss-rate and contributing to the reduction of \ac{CRC} related death.

Encoder-decoder based networks are widely used for automatic polyp segmentation~\cite{dong2021polyp,fan2020pranet,jha2020doubleu,rahman2023medical,sanderson2022fcn,tang2022duat,tomar2022tganet,zhang2021transfuse,zhao2021automatic}. Dong et al.~\cite{dong2021polyp} proposed Polyp-PVT that is based on pyramid vision transformer for automatic polyp segmentation. Wang et al.~\cite{wang2022stepwise} proposed SSFormer, which uses a pyramid Transformer encoder and progressive locality decoder to improve the performance of polyp segmentation. Encoder-decoder based methods achieved improved accuracy on regular or large-sized polyps. However, most methods are developed on white light imaging modality and are tested on an in-distribution dataset (tested on the same center). They readily fail on \ac{OOD} datasets such as small, diminutive, flat, sessile, or partially visible polyps and in the presence of camouflage and noisy images. Moreover, the complex morphological structures, indistinct boundaries between polyps and mucosa, and varying sizes make polyp segmentation more challenging. Therefore, there is an urgent need to develop a more generalizable and robust polyp segmentation method. 

We hypothesize that a prototype based segmentation can be a strong approach for OOD generalization because it relies on the creation of prototypes that capture the essential features of each class of objects in the images. These prototypes can be used to identify and segment objects even in images that differ significantly from those used during training.
An algorithm performing well on \ac{OOD} dataset could prevent models from making inaccurate diagnoses or treatment planning. Failure to do so can result in false positives or false negatives leading to misdiagnosis, which might have adverse consequences for the patient. Therefore, the model must perform well on \ac{OOD} datasets to be useful in clinical settings. In this work, we evaluate the proposed model on three datasets collected in different countries captured under different conditions and image enhancement techniques, including \ac{OOD} datasets to study the generalization ability of the model, which is critical for the development of \ac{CADx} system.  


\textbf{Summary of our contributions:}  \textbf{(1)} We propose to develop a  \textit{prototype learning} algorithm for medical image segmentation. To generate multiple prototypes, we design a Prototype Generation Module (PGM) by capturing the underlying data distribution. These prototypes handle variability and exhibit strong generalization capabilities when applied to novel data, \textbf{(2)} We present a \textit{Coarse Mask Generation Module (CMGM)} to improve the accuracy, efficiency, and generalization capabilities of the polyp segmentation network, \textbf{(3)} We encourage our network to operate on multi-scale fashion through an \textit{Encoder Feature Fusion Module (EFFM)},  \textbf{(4)} We devise a \textit{Prototype Mask Generation Module (PMGM)} to generate a final prototype mask using the prototypes generated by the PGM and the output feature map of the EFFM, and \textbf{(5)} we conduct a thorough analysis and evaluation on multi-center datasets and perform an \ac{OOD} generalization test on different datasets. Our experiments and results show that PrototypeLab outperforms $16$ SOTA medical image segmentation methods on three publicly available polyp datasets in terms of accuracy and efficiency.
\section{Method}
\begin{figure} [!t]
    \centering
    \includegraphics[width=\textwidth]{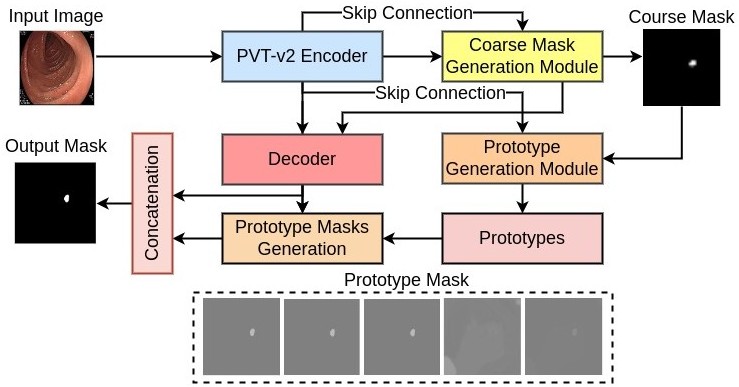}
    \caption{The overall architecture of the proposed PrototypeLab. The input image is fed to the PVT-encoder to generate a coarse mask, which is combined with encoder features in the prototype generation module to generate various prototypes. Subsequently, the decoder is employed, which produces a feature map that is used to create the prototype masks. Finally, the `output mask' is generated.}
    \label{fig:prototypelab}
\end{figure}

{\label{section:method}}
The proposed PrototypeLab is a new  image segmentation architecture with integrated \textit{prototype learning},  generating high-quality segmentation masks.  PrototypeLab consists of five key components: Pyramid Vision Transformer (PVT) encoder, Coarse Mask Generation Module (CMGM), Prototype Generation Module (PGM), Decoder, and Prototype Mask Generation Module (PMGM) (Figure~\ref{fig:prototypelab}). The CMGM, PGM, and PMGM are integral components that effectively contribute to the prototype learning process and collectively enhance the overall performance of the proposed architecture.

\subsection{Pyramid Vision Transformer (PVT) Encoder}
In PrototypeLab, the pyramid vision transformer (PVT) is used as a pre-trained encoder, extracting multi-scale features from an input image. An input image $I \in \mathbb{R}^{H\times W\times 3}$ is fed to the PVT-encoder to obtain four distinct feature maps $X_i \in \mathbb{R}^{\frac{H}{2^{i+1}} \times \frac{W}{2^{i+1}} \times C_i}$, where  $i \in \{1, 2, 3, 4\}$ and $C_i \in \{64, 128, 320, 512\}$. These feature maps are then  used in the other modules of PrototypeLab.
\subsection{Coarse Mask Generation Module (CMGM)}

The CMGM module generates a loose mask using features learned by the PVT encoder. It begins with an upsampling of $X_4$ to double its spatial dimensions, followed by concatenation with $X_3$ and passing through $3\times3$ Conv2D-BN-ReLU layers. The module incorporates a novel \textit{Large Kernel Dilated Convolution (LKDC)} block (Supp. Figure 1) consisting of two parallel convolution layers with large kernel size of $7\times7$ and $13\times13$ factorized into $\{7 \times 1\}\{1 \times 7\}$ and $\{13 \times 1\}\{1 \times 13\}$ respectively. The outputs of these layers are concatenated and fed into parallel dilated convolution layers with dilation rates of $r = {1, 2, 4}$. The utilization of large kernel size along dilated convolutions enhance the receptive field for capturing contextual information and handling objects of varying scales. The output is processed through a $1\times1$ Conv2D-BN-ReLU layer and upsampled by a factor of two. A final $1\times1$ convolution layer with sigmoid activation generates the coarse mask. The LKDC block enhances accuracy, efficiency, and generalization capabilities of the segmentation network.


\subsection{Prototype Generation Module (PGM)}

The PGM module utilizes PVT encoder features and a coarse mask to generate multiple prototypes. It applies $3\times3$ Conv2D-BN-ReLU layers to the PVT-encoder feature $X_i$. The feature maps are then element-wise multiplied with the coarse mask, followed by mask average pooling. This process yields four prototypes $P = \{p_1, p_2, p_3, p_4\}$ from $X_i = \{X_1, X_2, X_3, X_4\}$. {Learning multiple prototypes allows for a richer representation of diverse visual patterns. By utilizing different features from the PVT encoder, we can capture a wider range of information and enhance the model's ability to recognize various objects and scenarios. This approach improves the robustness of the system by increasing its adaptability to different input conditions, resulting in more accurate and reliable predictions.} The number of prototypes can be adjusted based on the problem complexity.

For multi-scale feature fusion of PVT encoder features, we introduce the Encoder Feature Fusion Module (EFFM) (Supp. Figure 1). It involves upsampling $X_4$ and concatenating it with $X_3$, followed by $3\times3$ Conv2D-BN-ReLU layers. This process is repeated for upsampling and concatenation with $X_2$, and again with $X_1$. Four parallel convolution layers are used, including a $1\times1$ convolution layer and three factorized convolution layers with dilated convolutions. The outputs of these layers are concatenated and passed through a final $1\times1$ convolution layer to generate the final prototype $p_5$.

\subsection{Decoder}

In the decoder, the upsampled feature map from the CMGM is first concatenated with $X_2$ and passed through a residual block. Then, the feature map is upsampled by a factor of two and concatenated with $X_1$, followed by another residual block. Subsequently, we upsample the feature map to increase the spatial dimensions by a factor of four, and then concatenate it with the original input image. This is followed by another residual block. Finally, the output of this block is used to generate prototype masks.
\vspace{-1mm}
\subsection{Prototype Mask Generation Module and Final Mask}

The PMGM utilizes cosine similarity to generate five sets of masks by comparing the decoder's output feature map with multiple prototypes from the PGM. To create the final segmentation mask, we concatenate the decoder's output feature map with the prototype masks. This concatenated feature map undergoes a residual block, followed by a $1\times1$ convolution and a sigmoid activation function. {The proposed framework effectively incorporates the CMGM, PGM, decoder, and PMGM to generate accurate and efficient polyp segmentation masks.}

\section{Experiments and Results}
\begin{table}[t!]
\footnotesize
\centering
\caption{Result of models trained and tested on BKAI-IGH~\cite{lan2021neounet}. `\textcolor{red}{Red}', `\textcolor{green}{Green}' and `\textcolor{blue}{Blue}' colors represent the highest, second highest and third highest scores.}
 \begin{tabular} {l|c|c|c|c|c|c|c}
\toprule
\textbf{Method}  &\textbf{Publication} &\textbf{mDSC}  & \textbf{mIoU}  &\textbf{Recall}& \textbf{Precision} &\textbf{F2} &\textbf{HD} \\ 
\hline
U-Net~\cite{ronneberger2015u} &MICCAI 2015	&0.8286 &0.7599		&0.8295	&0.8999	&0.8264	&3.17\\
DeepLabV3+~\cite{chen2018encoder} &ECCV 2018 &0.8938 &0.8314		&0.8870	&0.9333	&0.8882	&2.90\\
PraNet~\cite{fan2020pranet} &MICCAI 2021	&0.8904 &0.8264		&0.8901	&0.9247	&0.8885	&2.94\\
MSNet~\cite{zhao2021automatic}&MICCAI 2021	&0.9013 &0.8402		&0.8868	&\textcolor{blue}{0.9423} &0.8913	&2.85\\
TransFuse-S~\cite{zhang2021transfuse}&MICCAI 2021	&0.8599 &0.7819		&0.8531	&0.9075	&0.8530	&3.04\\
TransFuse-L~\cite{zhang2021transfuse}&MICCAI 2021	&0.8747 &0.8105		&0.8736	&0.9235	&0.8723	&2.96\\
Polyp-PVT~\cite{dong2021polyp}&ArXiv 2021		&0.8995 &0.8379		&0.9016	&0.9238	&0.8986	&2.88\\
UACANet~\cite{kim2021uacanet} &ACMMM 2021	&0.8945 &0.8275  &0.8870 &0.9297 &0.8882 &2.86	\\ 
DuAT~\cite{tang2022duat} &Arxiv 2022	&\textcolor{green}{0.9140} &\textcolor{green}{0.8563}		&0.9038	&\textcolor{green}{0.9437}	&\textcolor{blue}{0.9066}	&\textcolor{blue}{2.77}\\
CaraNet~\cite{lou2022caranet}	&MIIP 2022	&0.8962 &0.8329		&0.8939	&0.9273	&0.8937	&2.91\\
SSFormer-S~\cite{wang2022stepwise}&MICCAI 2022	&0.9111	&0.8527		&\textcolor{blue}{0.9043}	&0.9391	&0.9060	&2.81\\
SSFormer-L~\cite{wang2022stepwise}	&MICCAI 2022&\textcolor{blue}{0.9124}	&0.8508 &0.9005	&0.9400	&0.9041	&\textcolor{green}{2.74} \\
UNeXt~\cite{valanarasu2022unext}&MICCAI 2022	&0.4758 &0.3797  &0.5814 &0.5820 &0.5132 &4.49	\\
LDNet~\cite{zhang2022lesion}&MICCAI 2022	&0.8927	&0.8254		&0.8867	&0.9153	&0.8874	&2.94 \\
TGANet~\cite{tomar2022tganet}&MICCAI 2022	&0.9023	&0.8409		&0.9025	&0.9208	&0.9002	&2.84\\
PVT-Cascade~\cite{rahman2023medical}&WACV 2023	&0.9123		&\textcolor{blue}{0.8534}	&\textcolor{red}{\textbf{0.9223}}	&0.9212	&\textcolor{green}{0.9167}	&2.81\\
\textbf{PrototypeLab} & &\textcolor{red}{\textbf{0.9243}} &\textcolor{red}{\textbf{0.8744}}	 & \textcolor{green}{0.9194} &\textcolor{red}{\textbf{0.9494}}&\textcolor{red}{\textbf{0.9202}}	&\textcolor{red}{\textbf{2.70}}\\
\bottomrule
\end{tabular}
\label{tab:resultsbkai}
\vspace{-5mm}
\end{table}

\begin{table}[t!]
\footnotesize
\centering
\caption{Quantitative results {of the model trained and tested on} Kvasir-SEG~\cite{jha2020kvasir}.}
 \begin{tabular} {l|c|c|c|c|c|c}
\toprule
\textbf{Method}  &\textbf{mDSC}  & \textbf{mIoU}   &\textbf{Recall}& \textbf{Precision} &\textbf{F2} &\textbf{HD} \\ 
\hline

U-Net~\cite{ronneberger2015u}&0.8264 &0.7472 &0.8503	&0.8703	&0.8352	&4.57\\
DeepLabV3+~\cite{chen2018encoder}&0.8837 &0.8172 &0.9014	&0.9028	&0.8900	&4.10\\
PraNet~\cite{fan2020pranet}	&0.8943 &0.8296	&0.9060	&0.9126 &0.8976	&4.00\\
MSNet~\cite{zhao2021automatic}&0.8859	&0.8217	&0.9006	&0.9110	&0.8901	&4.01\\
TransFuse-S~\cite{zhang2021transfuse}&0.8780 &0.8079	&0.8898	&0.9090	&0.8813	&4.09 \\
TransFuse-L~\cite{zhang2021transfuse}&0.8768 &0.8115	&0.8842	&\textcolor{green}{0.9198}	&0.8771	&4.05\\
Polyp-PVT~\cite{dong2021polyp}&0.8960	&0.8328		&\textcolor{red}{\textbf{0.9440}}	&0.8811	&\textcolor{green}{0.9164}	&3.91\\
UACANet~\cite{kim2021uacanet} &0.8835 &0.8133  &0.9085 &0.8947 &0.8937 &4.19	\\
DuAT~\cite{tang2022duat}&0.8903	&0.8294		&0.9186	&0.9019	&0.8999	&\textcolor{blue}{3.87}\\
CaraNet~\cite{lou2022caranet}&0.8707	&0.7958		&0.9203	&0.8621	&0.8935	&4.11\\
SSFormer-S~\cite{wang2022stepwise}&\textcolor{blue}{0.8994}	&\textcolor{blue}{0.8363}		&0.9194	&0.9086	&0.9076	&3.88\\
SSFormer-L~\cite{wang2022stepwise}&\textcolor{green}{0.9060}	&\textcolor{green}{0.8500}	&0.9213	&\textcolor{red}{\textbf{0.9199}}	&\textcolor{blue}{0.9107}	&\textcolor{green}{3.77}\\
UNeXt~\cite{valanarasu2022unext} &0.7318 &0.6284  &0.7840 &0.7656 &0.7507 &5.18	\\
LDNet~\cite{zhang2022lesion}	&0.8881 &0.8208		&0.9063	&0.9046	&0.8946	&4.09\\
TGANet~\cite{tomar2022tganet}	&0.8982 &0.8330		&0.9132	&0.9123	&0.9029	&3.96	\\
PVT-Cascade~\cite{rahman2023medical} &0.8950 	&0.8329		&\textcolor{green}{0.9355}	&0.8888	&0.9103	&3.90\\
\textbf{PrototypeLab}	&\textcolor{red}{\textbf{0.9086}} &\textcolor{red}{\textbf{0.8544}} &\textcolor{blue}{0.9344}	& \textcolor{blue}{0.9136}	&\textcolor{red}{\textbf{0.9194}}	&\textcolor{red}{\textbf{3.71}}\\
\bottomrule
\end{tabular}
\label{tab:resultsKvasir-seg}
\end{table}

\begin{table}[t!]
\footnotesize
\centering
\caption{Result of {models trained on BKAI-IGH}~\cite{lan2021neounet} and tested on PolypGen~\cite{ali2023multi}.}
 \begin{tabular} {l|c|c|c|c|c|c}
\toprule
\textbf{Method} &\textbf{mDSC}  & \textbf{mIoU}  &\textbf{Recall}& \textbf{Precision} &\textbf{F2} &\textbf{HD}\\ 
\hline
U-Net~\cite{ronneberger2015u}&0.5841	&0.5102		&0.6142	&0.7746	&0.5739	&4.36\\
DeepLabV3+~\cite{chen2018encoder}&0.6757	&0.6051		&0.7074	&0.8237	&0.6732	&4.03\\
PraNet~\cite{fan2020pranet}	&0.7330 &0.6659		&0.7825	&0.8182	&0.7391	&3.71\\
MSNet~\cite{zhao2021automatic}&0.6777	&0.6133		&0.6811	&\textcolor{red}{\textbf{0.8881}}	&0.6657	&4.06\\
TransFuse-S~\cite{zhang2021transfuse}&0.6510	&0.5720		&0.6894	&0.7952	&0.6416	&4.04\\
TransFuse-L~\cite{zhang2021transfuse}&0.6592	&0.5881		&0.6792	&0.8289	&0.6487	&4.10\\
Polyp-PVT~\cite{dong2021polyp}&\textcolor{blue}{0.7421}  &\textcolor{green}{0.6746} &0.7717	&0.8494	&\textcolor{blue}{0.7358}	&3.67\\
UACANet~\cite{kim2021uacanet}&0.7063 &0.6404  &0.7265 &\textcolor{green}{0.8519} &0.7016 &3.87 \\ 
DuAT~\cite{tang2022duat}&0.7225	&0.6553		&0.7710	&0.8081	&0.7204	&3.73\\
CaraNet~\cite{lou2022caranet}&0.6977	&0.6329		&0.7035	&\textcolor{blue}{0.8830} &0.6873	&3.99\\
SSFormer-S~\cite{wang2022stepwise}&0.7332 &0.6664		&0.7672	&0.8386	&0.7288	&3.72\\
SSFormer-L~\cite{wang2022stepwise}&\textcolor{green}{0.7426}	&\textcolor{blue}{0.6732}	&\textcolor{green}{0.7901}	&0.8209	&\textcolor{green}{0.7418}	&\textcolor{green}{3.60}\\
UNeXt~\cite{valanarasu2022unext}&0.3484 &0.2669  &0.4519 &0.4834 &0.3492 &5.09\\
LDNet~\cite{zhang2022lesion}&0.6922 &0.6193		&0.7389	&0.8013	&0.6836	&3.83\\
TGANet~\cite{tomar2022tganet}&0.6925	&0.6206		&0.7466	&0.7833	&0.6891	&3.79\\
PVT-Cascade~\cite{rahman2023medical}&0.7271	&0.6572		&\textcolor{red}{\textbf{0.8154}} &0.7672	&\textcolor{blue}{0.7358}	&\textcolor{red}{\textbf{3.58}}\\
\textbf{PrototypeLab} &\textcolor{red}{\textbf{0.7583}} &\textcolor{red}{\textbf{0.6957}}	 &\textcolor{blue}{0.7897}	&0.8456 &\textcolor{red}{\textbf{0.7563}} &\textcolor{blue}{3.68}\\
\bottomrule
\end{tabular}
\label{tab:resultsBKAI-Polypgen}
\end{table}

\textbf{Datasets:} We use BKAI-IGH~\cite{lan2021neounet}, Kvasir-SEG~\cite{jha2020kvasir}, and PolypGen~\cite{ali2023multi} dataset for experimentation. The BKAI-IGH consists of 1000 images with the corresponding ground truth. The dataset was collected from two medical centers in Vietnam. It contains images captured using WLI, FICE, BLI and LCI. To train all the algorithms, we used 800 images for training, 100 for validation, and 100 for testing. Similarly, Kvasir-SEG consists of 1000 images collected from four hospitals in Norway. We have used 880 in the training set and {rest} 120 images in the {validation and} test set. Moreover, we use PolypGen~\cite{ali2023multi} dataset collected from six medical centers in \textit{Norway, United Kingdom, France, Italy and Egypt} as \ac{OOD} dataset. PolypGen is collected from multiple hospitals that cover different clinical patient populations and modalities, which makes it diverse and useful for generalizability tests. 

\noindent\textbf{Experimental Setup:} We have trained all the models on NVIDIA GeForce RTX 3090 GPU. All the images are first resized to $256 \times 256$ pixels. The training images are followed  by simple data augmentation strategies, which includes random rotation, vertical flipping, horizontal flipping, and coarse dropout, are used to improve generalization and prevent overfitting. All models are trained on a similar hyperparameters configuration with a learning rate of $1e^{-4}$, batch size of 16, and an ADAM optimizer. We use a combination of binary cross-entropy and dice loss with equal weights as a loss function. In addition, we use an early stopping and \textit{ReduceLROnPlateau} to avoid overfitting.

\begin{figure} [!t]
    \centering
    \includegraphics[width=0.8\textwidth]{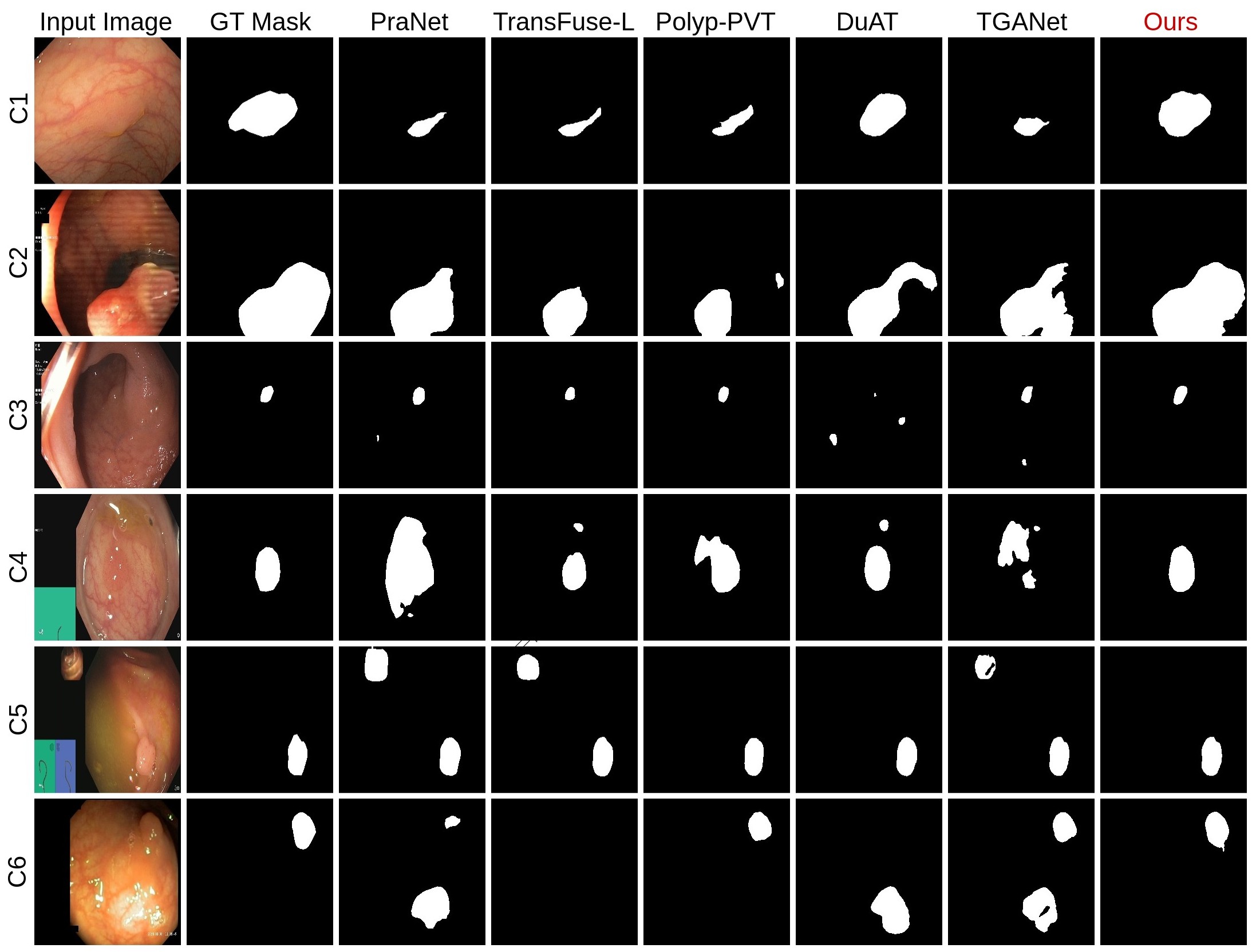}
    \caption{Qualitative results of models trained BKAI-IGH and tested on PolypGen. It can be observed that PrototypeLab produces a more accurate segmentation map in all the centers from C1 to C6.}
    \label{fig:qualitative}
\end{figure}

\begin{figure} [!h]
    \centering
    \includegraphics[width=0.7\textwidth]{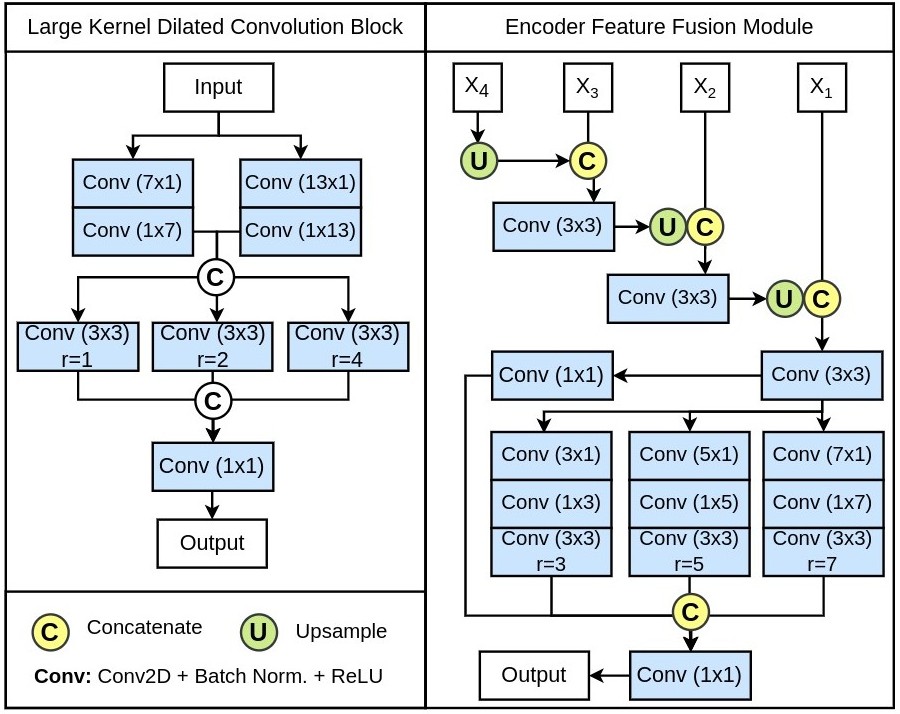}
    \caption{The diagram of the Large Kernel Dilated Convolution block and Encoder Feature Fusion Module.}
    \label{fig:lkdc_effm}
\end{figure}

\begin{figure} [!h]
    \centering
    \includegraphics[width=0.8\textwidth]{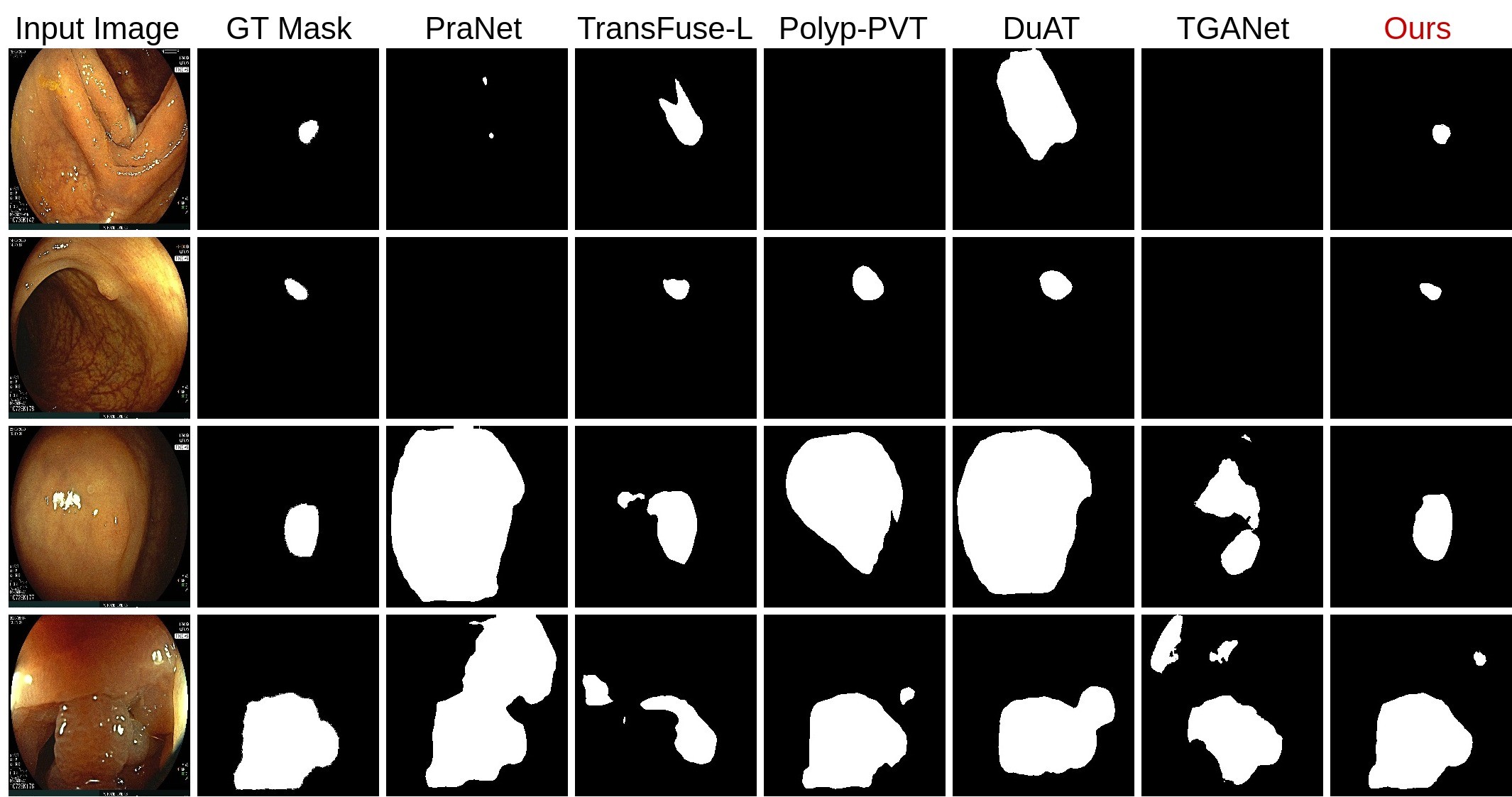}
    \caption{Qualitative results of the models trained on Kvasir-SEG and tested on BKAI-IGH.}
    \label{fig:my_label}
\end{figure}

\begin{table}[!t]
\centering
\footnotesize
\caption{The study provides model parameters, flops, and FPS for both SOTA methods and proposed PrototypeLab. `Red', `Green', and `Blue' represent the best, second best and third best scores.}
\begin{tabular}{l|c|c|c|c|c} 
\toprule
\textbf{Method} &\shortstack{\textbf{Publication}\\ \textbf{Venue}} &\textbf{Backbone} & \shortstack{\textbf{Param.}\\ (\textbf{Million)}} &\shortstack{\textbf{Flops}\\\textbf{(GMac)}} & \textbf{FPS}\\ \midrule 

TransFuse-S & MICCAI 2021 &ResNet34 + DeiT-S &26.35 &11.5 & 34.76\\
TransFuse-L& MICCAI 2021 &ResNet50 + DeiT-B &143.74 &82.71 &37.6\\
Polyp-PVT&ArXiv 2021 &PVTv2-B2 &\textcolor{blue}{25.11} &\textcolor{green}{5.30} & 47.54 \\

UACANet &ACMMM 2021 & Res2Net50 &69.16 &31.51	&26.04 \\ 
DuAT& Arxiv 2022 &PVTv2-B2 &\textcolor{green}{24.97} &\textcolor{red}{\textbf{5.24}} & 44.56\\
CaraNet & MIIP 2022 &Res2Net101 &46.64	&11.48 &20.67 \\
SSFormer-S& MICCAI 2022 &MiT-PLD-B2 &29.57 &10.1 &49.11\\
SSFormer-L&MICCAI 2022 &MiT-PLD-B4 &66.22 &17.28 &28.21\\
UNeXt	&MICCAI 2022 & -	&\textcolor{red}{\textbf{1.47}} &0.5695 &\textcolor{green}{88.01} \\
LDNet& MICCAI 2022 &Res2Net50 &33.38	&33.14  &31.14\\
TGANet & MICCAI 2022 &ResNet50 &19.84	&41.88 &26.21 \\
PVT-Cascade & WACV 2023 &PVTv2-B2 &35.27	&\textcolor{blue}{8.15} &39.09\\
\textbf{PrototypeLab} & - &PVTv2-B2 &51.34 &174.87 &23.85 \\
\bottomrule
\end{tabular}
\label{algorithm_complexity}
\end{table}

\begin{table}[t!]
\footnotesize
\centering
\caption{Results of models trained on Kvasir-SEG and tested on PolypGen. `Red', `Green', and `Blue' represent the best, second best and third best scores.}
 \begin{tabular} {l|c|c|c|c|c|c}
\toprule
\textbf{Method} &\textbf{mDSC}  & \textbf{mIoU}   &\textbf{Recall}& \textbf{Precision}  &\textbf{F2} &\textbf{HD}\\ 
\hline
TransFuse-S	&0.6956	&0.6229	&0.8079	&0.7481	&0.7180	&3.66\\
TransFuse-L&0.7273	&0.6621		&0.8060	&0.7822	&0.7404	&3.53\\
Polyp-PVT&\textcolor{blue}{0.7424}	&\textcolor{blue}{0.6763}		&0.8369	&0.7985	&0.7557	&3.44\\
UACANet&0.7185 &0.6517  &0.7972 &\textcolor{blue}{0.8007} &0.7288 &3.61\\ 
DuAT&0.7332	&0.6654	&\textcolor{green}{0.8664}	&0.7342	&0.7609	&3.48\\
CaraNet&0.6976 &0.6240	&0.8264	&0.7419	&0.7285	&3.68\\
SSFormer-S&0.7389	&0.6734	&0.8625	&0.7496	&0.7634	&\textcolor{blue}{3.43}\\
SSFormer-L&\textcolor{green}{0.7556}	&\textcolor{green}{0.6926}	&0.8530	&0.7910 &\textcolor{green}{0.7703}	&\textcolor{green}{3.37}\\
UNeXt&0.4552  &0.3761  &0.6135 &0.5600 &0.4805 &4.55 \\
LDNet&0.7273	&0.6604		&0.8336	&0.7685	&0.7462	&3.53\\
TGANet&0.7030	&0.6386		&0.8030	&0.7654	&0.7177	&3.59\\
PVT-Cascade&0.7151	&0.6460		&\textcolor{blue}{0.8651} &0.7108	&0.7396	&3.48\\
\textbf{PrototypeLab}&\textcolor{red}{\textbf{0.7560}}	&\textcolor{red}{\textbf{0.6966}}	&0.8603	&0.7846 &\textcolor{red}{\textbf{0.7745}}	&\textcolor{red}{\textbf{3.35}}\\
\bottomrule
\end{tabular}
\label{tab:results}
\end{table}

\begin{table*}[!t]
\footnotesize
\centering
\caption{Ablation study of PrototypeLab on OOD dataset. The methods are trained on Kvasir-SEG dataset and tested on PolypGen dataset.}
 \begin{tabular} {l|l|c|c|c}
\toprule
\textbf{No}  &\textbf{Method} & \textbf{mDSC}  &\textbf{mIoU}  &\textbf{HD}\\
\midrule

\#1 &Baseline (PVT-encoder + Decoder) &0.7503 &0.6879 &\textcolor{green}{\textbf{3.36}} \\

\#2 &Baseline + CMGM &\textcolor{red}{\textbf{0.7585}} &\textcolor{green}{\textbf{0.6956}} &\textcolor{green}{\textbf{3.36}}\\

\#3 &Baseline + (CMGM w/o LKDC) + PGM + PMGM &0.7234 &0.6566 &3.49\\

\#4 &Baseline + CMGM + (PGM w/o EFFM) + PMGM &\textcolor{blue}{\textbf{0.7538}} &0.6909 &\textcolor{green}{\textbf{3.36}}\\

\#5 &Baseline + (CMGM w/o LKDC) + (PGM w/o EFFM) + PMGM  &0.7521 &\textcolor{blue}{\textbf{0.6910}} &\textcolor{blue}{\textbf{3.39}}\\

\#6 &Baseline + CMGM + PGM + PMGM (PrototypeLab)  &\textcolor{green}{\textbf{0.7560}} &\textcolor{red}{\textbf{0.6966}} &\textcolor{red}{\textbf{3.35}} \\

\bottomrule
\end{tabular}
\label{ablation}
\end{table*}

\paragraph{\textbf{Results on BKAI-IGH dataset:}} Table~\ref{tab:resultsbkai} shows the results of all the models on BKAI-IGH dataset. The table demonstrates the superiority of PrototypeLab  with a high \ac{DSC} of 0.9243, \ac{mIoU} of 0.8744, a high recall of 0.9194, a precision of 0.9494,  low \ac{HD} of 2.70. It outperforms 16 \ac{SOTA} methods.  DuAT~\cite{tang2022duat} and SSFormer-L~\cite{wang2022stepwise} are the most competitive network to our network, where our network still outperforms DuAT and FCBFormer by 1.03\%  and 1.19\% in \ac{DSC} respectively. These results suggest that PrototypeLab is highly effective in segmenting polyps on different endoscopic imaging techniques such as WLI, BLI, FCI, and FICE.   
\paragraph{\textbf{Results on Kvasir-SEG dataset:}} Table~\ref{tab:resultsKvasir-seg} shows the results of all the models on Kvasir-SEG dataset. PrototypeLab obtains a high \ac{DSC} of 0.9086, \ac{mIoU} of 0.8544, recall of 0.9344, precision of  0.9136, and low \ac{HD} of 3.71. The most competitive network to PrototypeLab is SSFormer-L~\cite{wang2022stepwise}.  Our model surpasses by SSFormer-L by 0.26\% in \ac{DSC}, 0.44\% in \ac{mIoU}, 0.06 in \ac{HD} metrics. Thus, PrototypeLab surpasses all \ac{SOTA} in both overlap based metrics and distance based metrics. 

\paragraph{\textbf{Results of BKAI-IGH models tested on PolypGen:}} Table~\ref{tab:resultsBKAI-Polypgen} shows results of the model trained on BKAI-IGH and tested on PolypGen (all centers combined). PrototypeLab obtains the highest \ac{DSC}, \ac{mIoU}, and F2 of 0.7583, 0.6957 and 0.7563, respectively. SSFormer-L and Polyp-PVT are the most competitive network, where PrototypeLab outperforms both networks by 1.57\% and 1.62\%, respectively. Although PVT-Cascade has the least \ac{HD}, our results is very competitive. We have similar findings when the model is trained on Kvasir-SEG and tested on PolypGen (Supp. Table 2) which suggests that PrototypeLab is more effective in handling \ac{OOD} dataset.

Figure~\ref{fig:qualitative} and \textit{Supp. Figure 2} shows the qualitative results comparison of different models on diminutive polyp, flat polyp and regular polyp. The most competitive network, DuAT and Polyp-PVT exhibit over-segmentation or under-segmentation on different scenarios. However, PrototypeLab can segment more accurately on diminutive, flat and noisy images as compared to the \ac{SOTA} baselines. \textit{Supp. Table 1} shows the number of parameters, flops and processing speed. The Table shows that PrototypeLab has 51.34 Million parameters and 174. 87 GMac Flops with a processing speed of 23.85. Although the processing speed is close to near real-time ($\approx 30 fps$), our architecture is more accurate, which is essential in clinical settings for early diagnosis and treatment. Therefore, the trade-off between speed and increased accuracy can be compensated. 



\paragraph{\textbf{Ablation study:}}
Table~\ref{ablation} shows the ablation study of the PrototypeLab on the Kvasir-SEG. The results show that the baseline (\#1) obtains \ac{DSC} of 0.8971, \ac{mIoU} of 0.8399, and \ac{HD} of 3.88. In setting \#2, a slight performance improvement was observed. This is because the masks generated by the CMGM were not used in the decoder, but were used by the PGM to generate multiple prototypes. These prototypes were then utilized by the PMGM to generate multiple prototype mask in setting \#6, which explains the limited performance improvement when comparing setting \#2 to setting \#1.To demonstrate the impact of LKDC block and EFFM, we have conducted three experiments in setting \#3, \#4 and \#5, where we can observed a drop in performance when compared with setting \#6. Specifically, when both the LKDC block and EFFM were removed in setting \#5, a 0.71\% decrease in \ac{DSC}, a 0.80\% decrease in \ac{mIoU}, and a 0.11\% increase in \ac{HD} were observed compared to setting \#6 The Table shows that the baseline is improved by adding CMGM and further improved by adding PGM + PMGM. PrototypeLab (\#6) offers an improvement of 1.15\% in \ac{DSC}, 1.45\% in \ac{mIoU} and 0.17\% in \ac{HD} {when compared with baseline}.

\begin{table*}[!t]
\scriptsize
\centering
\caption{Ablation study of PrototypeLab on Kvasir-SEG. Here, BL = Baseline.}
\begin{tabular} {l|l|c|c|c|c}
\toprule
\textbf{No}  &\textbf{Method} & \textbf{mDSC}  &\textbf{mIoU}  &\textbf{Recall} &\textbf{HD}\\
\midrule
\#1 &BL (PVT-encoder + Decoder) &0.8971 &0.8399 &0.9203  &3.88\\
\#2 &BL + CMGM  &0.8971 &0.8402 &0.9199  &3.84\\

{\#3} &{BL} + {(CMGM w/o LKDC)} + {PGM} +  {PMGM} & {0.8935} &{0.8332} &{0.9282} &{4.02}	\\

{\#4} &{BL} + {CMGM} + {(PGM w/o EFFM)} +  {PMGM}  &{0.9012} &{0.8440} &{0.9284} &{3.82}	\\

{\#5} &{BL} + {(CMGM w/o LKDC)} +  {\shortstack{(PGM w/o EFFM)} + {PMGM}} &{0.9015} &{0.8464} &{0.9228} &{3.82}	\\

\#6 & {BL + CMGM + PGM + \shortstack{PMGM  (PrototypeLab)}} &\textbf{0.9086} &\textbf{0.8544} &\textbf{0.9344} &\textbf{3.71} \\
\bottomrule
\end{tabular}
\label{ablation}
\end{table*}
\section{Conclusion}
We propose a prototype learning based new segmentation model, called PrototypeLab, for in-distribution and out-of-distribution polyp segmentation. The use of prototypes in the proposed architecture helps in dealing with variability present in the medical images making the model more robust to inter-patient variations. It helps to perform well on diminutive, flat, partially visible, noisy images and camouflage properties of polyp. The proposed architecture obtains high \ac{DSC} of 0.9243, \ac{mIoU} of 0.8744, and low \ac{HD} of 2.70 on the BKAI-IGH dataset. Our extensive experiments revealed that PrototypeLab exhibits superior performance compared to 16 state-of-the-art (SOTA) methods across three distinct datasets, including notoriously difficult multi-center out-of-distribution (OOD) datasets.  In the future, we aim to develop PrototypeLabV2, by further optimizing speed and accuracy for mobile applications. 


\bibliographystyle{splncs04}
\bibliography{ref}
\end{document}